\documentclass[runningheads]{llncs}

\usepackage{eccv}

\usepackage{eccvabbrv}

\usepackage{graphicx}
\usepackage{booktabs}

\usepackage[accsupp]{axessibility}  

\usepackage[pagebackref,breaklinks,colorlinks,citecolor=eccvblue]{hyperref}
\usepackage{hyperref}

\usepackage{orcidlink}

\begin{document}

\title{Automatic Generation of Fashion Images using
Prompting in Generative Machine Learning Models} 


\author{Georgia Argyrou\inst{1}\orcidlink{0009-0001-0642-0888} \and
Angeliki Dimitriou\inst{1}\orcidlink{0009−0001−5817−3794} \and
Maria Lymperaiou\inst{1}\orcidlink{0000-0001-9442-4186}
\and
Giorgos Filandrianos\inst{1}\orcidlink{0000-0002-7015-7746}
\and
Giorgos Stamou\inst{1}\orcidlink{0000-0003-1210-9874}}

\authorrunning{G. Argyrou et al.}

\institute{
Artificial Intelligence and Learning Systems Laboratory\\
School of Electrical and Computer Engineering \\
National Technical University of Athens \\
\email{argeorgiaa@gmail.com}, \email{\{angelikidim, marialymp, geofila\}@ails.ece.ntua.gr}, \email{gstam@cs.ntua.gr}}

\maketitle

\begin{abstract}
The advent of artificial intelligence has contributed in a groundbreaking transformation of the fashion industry, redefining creativity and innovation in unprecedented ways.
This work investigates methodologies for generating tailored fashion descriptions using two distinct Large Language Models and a Stable Diffusion model for fashion image creation. Emphasizing adaptability in AI-driven fashion creativity, we depart from traditional approaches and focus on prompting techniques, such as zero-shot and few-shot learning, as well as Chain-of-Thought (CoT), which results in a variety of colors and textures, enhancing the diversity of the outputs. Central to our methodology is Retrieval-Augmented Generation (RAG), enriching models with insights from fashion sources to ensure contemporary representations. Evaluation combines quantitative metrics such as CLIPscore with qualitative human judgment, highlighting strengths in creativity, coherence, and aesthetic appeal across diverse styles. Among the participants, RAG and few-shot learning techniques are preferred for their ability to produce more relevant and appealing fashion descriptions. Our code is provided at \href{https://github.com/georgiarg/AutoFashion}{https://github.com/georgiarg/AutoFashion}.
\end{abstract}

\section{Introduction}
\label{sec:intro}

Fashion is a dynamic and influential industry that not only reflects cultural and societal trends but also drives economic growth on a global scale. \cite{statista} The integration of artificial intelligence (AI) within the fashion industry has revolutionized various aspects, from design and manufacturing to marketing and retail\cite{liang}. AI technologies, such as computer vision and machine learning, have enhanced the efficiency and creativity of fashion processes, enabling innovations like personalized shopping experiences and predictive trend analysis. \cite{ecommerce, ai-driven-recommender, personalized, ayedi2023airecommendationenhancedcustomer}

As evidenced by prior research, the field of automatic fashion image generation has not yet achieved significant advancements \cite{zhuang2022chattodesignaiassistedpersonalized, fashion-design}. This paper seeks to address this gap by developing an automated process for generating fashion images that align with a specified style, match a particular occasion, and suit an individual wearer. Our primary goal is to investigate various prompting methodologies to generate fashion descriptions tailored to specific variables such as style, wearer's type, and occasion. We direct two distinct large language models (LLMs) to create these descriptions and then use a Stable Diffusion \cite{rombach2022highresolutionimagesynthesislatent} model to generate corresponding fashion images. Through this approach, we aim to advance the capabilities of automatic fashion image generation and contribute to the evolving intersection of fashion and technology.

We shift from the traditional "pre-train, fine-tune" to a "pre-train, prompt, and predict" strategy, relying solely on prompting to guide the LLMs \cite{DBLP:journals/corr/abs-2107-13586}. To keep our models up-to-date with evolving fashion trends, we employ knowledge injection through Retrieval-Augmented Generation (RAG) \cite{gao2024retrievalaugmentedgenerationlargelanguage}. This involves incorporating insights from diverse sources such as fashion magazines, blogs, and relevant literature to ensure the relevance of the generated descriptions. Our findings are supported by extensive human surveys, which validate the quality of our generated images, while providing insights regarding the influence of each prompting method on the final image. We also employ traditional metrics such as the CLIPscore \cite{hessel2022clipscorereferencefreeevaluationmetric} to gauge text-image alignment.

Overall, our study highlights the importance of resource efficiency and demonstrates the effectiveness of LLMs in text generation tasks, particularly when combined with Stable Diffusion models for fashion image generation conditioned on text. 
In summary, this paper makes four major contributions:

\begin{itemize}
    \item To the best of our knowledge, we are the first to create an automated system for generating \textit{personalized} fashion outfit descriptions and images, tailored to specific styles, occasions, and individual wearer profiles.
    \item We create a dataset of fashion images accompanied with descriptions generated from two different LLMs.
    \item We guide LLMs using advanced prompting techniques and RAG, and develop hand-crafted templates for optimally generating fashion content.
    \item We collect and analyze human evaluation data to assess the quality and accuracy of the fashion content generated by our proposed method.
\end{itemize}

\section{Related Work}
Intelligent fashion is a challenging task due to the inherent variability in fashion items style and design, as well as the substantial semantic gap between computable low-level features and their corresponding high-level semantic concepts.\cite{cheng2021fashion}
As referenced in \cite{cheng2021fashion}, previous works relevant to intelligent fashion are divided into four distinct aspects. Fashion analysis encompasses attribute recognition, style learning, and popularity prediction, while fashion detection involves landmark detection, fashion parsing, and item retrieval. Fashion synthesis includes style transfer, pose transformation, and physical simulation. Also, fashion recommendation covers fashion compatibility, outfit matching, and hairstyle suggestion. 
However, research in fashion outfit description and fashion image generation is relatively limited. Fashion-Gen research team launches a challenge of text-to-image generation based on their dataset \cite{rostamzadeh2018fashiongen}. Other datasets \cite{jia2020fashionpedia, 7780493, DBLP:journals/corr/abs-1710-07346, yu2024quality, huang2023first, morelli2022dresscode, zheng2019modanet, DBLP:journals/corr/abs-1906-05750} have been created to facilitate fashion image generation or other fashion-related tasks. \cite{DBLP:journals/corr/abs-1710-07346} and \cite{9667042} present methods for redressing an individual in an image using a described fashion outfit.

\section{Methodology} 

Our proposed pipeline is illustrated in \cref{fig:model}. The input consists of variable triplets. In our experiments, we define two kinds of triplets: "style, occasion, gender" and "style, occasion, type." The "type" variable includes both the body type and the gender of the wearer. These triplets are driven to fill a custom prompt template, which varies depending on the prompting technique, to form a final prompt, which is then fed into an LLM; its output, which comprises the outfit description, serves as the input to an image generation model, ultimately producing the generated image.

The LLMs leveraged to produce the outfit description are Mistral-7B \cite{jiang2023mistral} and Falcon-7B\cite{almazrouei2023falcon}. For the image generation component, we used a Stable Diffusion model \cite{sdmodel, rombach2022highresolutionimagesynthesislatent}.
In order to produce desirable results, we do not use training or fine-tuning: varying prompting techniques are leveraged to guide the LLM, while knowledge injection manages to to keep it up-to-date.

\subsection{LLM Prompting}

\begin{figure}[h!]
  \centering
  \begin{subfigure}{1\linewidth}
  \centering
    \includegraphics[width=0.87\linewidth]{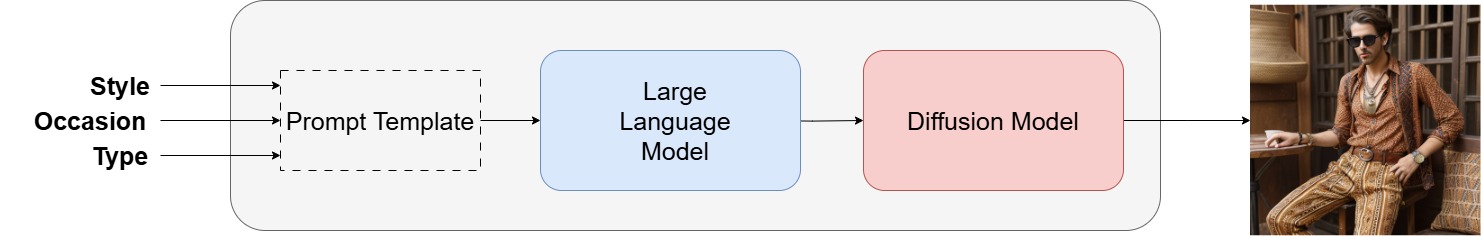}
    \caption{Zero-shot (ZS) learning}
    \label{fig:model}
  \end{subfigure}
  \hfill
  \begin{subfigure}{1\linewidth}
    \centering
    \includegraphics[width=0.87\linewidth]{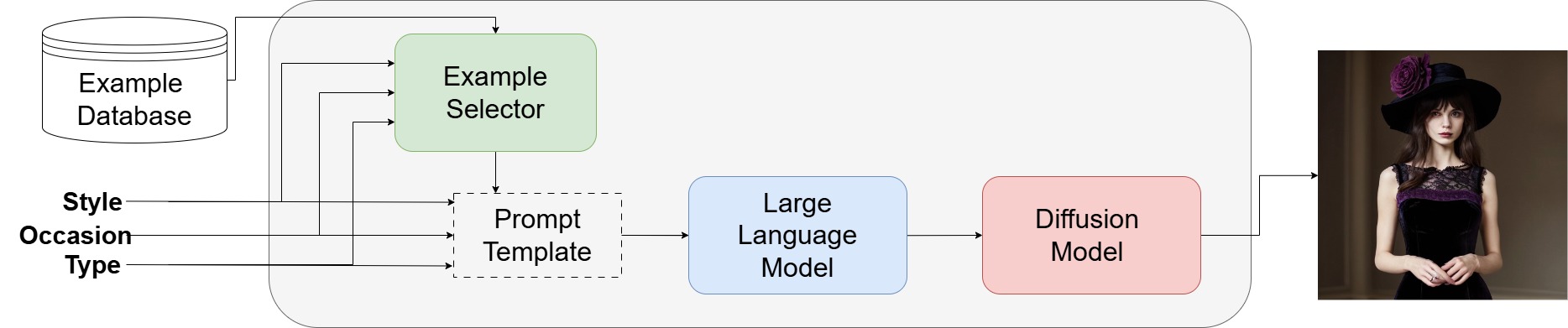}
    \caption{Few-shot (FS) learning}
    \label{fig:fs-model}
  \end{subfigure}
   \begin{subfigure}{1\linewidth}
     \centering
    \includegraphics[width=0.87\linewidth]{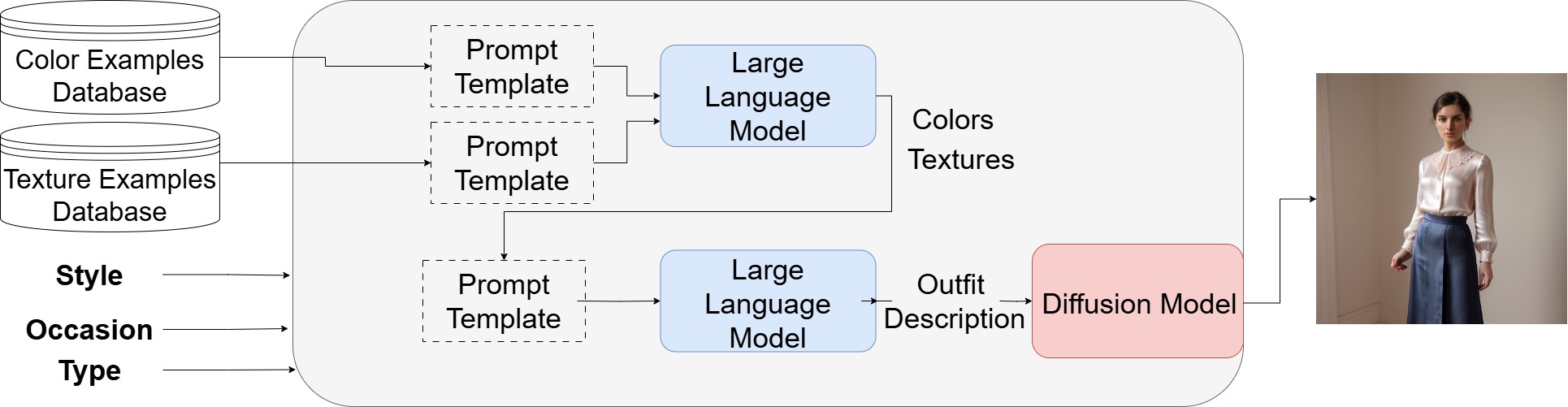}
    \caption{Chain-of-Thought (CoT)}
    \label{fig:cot-model}
  \end{subfigure}
    \begin{subfigure}{1\linewidth}
      \centering
    \includegraphics[width=0.87\linewidth]{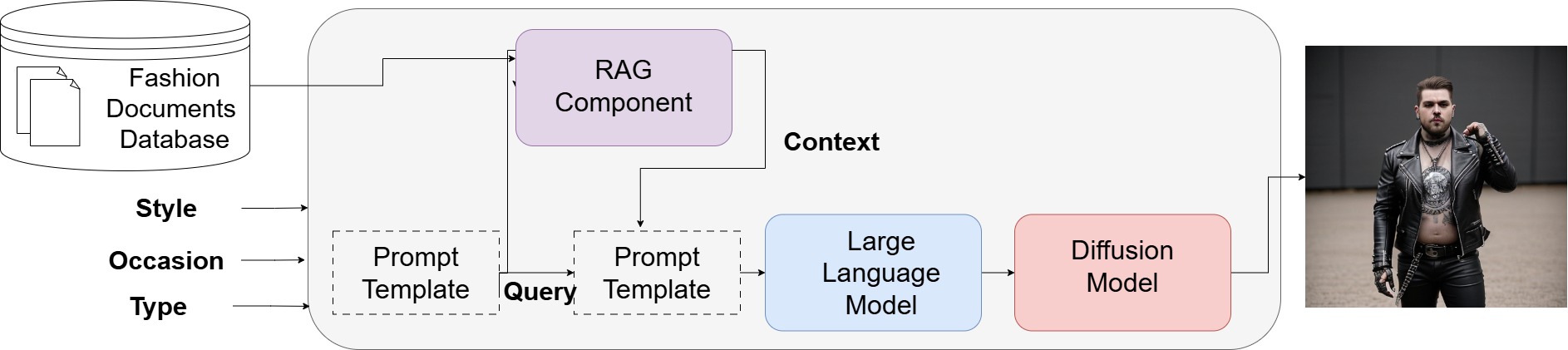}
    \caption{Retrieval-Augumented Generation (RAG)}
    \label{fig:rag-model}
  \end{subfigure}
  \caption{Fashion image generation pipelines using different prompting techniques and RAG to produce fashion outfit descriptions fed to a Stable Diffusion module.}
  \label{fig:models}
\end{figure}

For \textbf{zero-shot (ZS) learning} we utilize a simple template as shown in \cref{tab:templates} and follow exactly the process shown in \cref{fig:model}.

For \textbf{few-shot (FS) learning} we incorporate several examples as demonstrations. The prompt includes placeholders for the relevant examples that guide the LLM, as shown in \cref{tab:templates}. To ensure the examples are distinct and to clearly differentiate between the query and the answer, we formatted them as follows:
\begin{quote}
\begin{math}
\text{Question: [\textit{question}]} \newline
\text{Answer: [\textit{answer}]}
\end{math}
\end{quote}
In this format, \([question]\) contains queries aimed at generating an outfit description for a specific occasion, type, and style, while \([answer]\) provides detailed outfit descriptions that effectively respond to the given queries.
In addition to the process described in \cref{fig:model}, we incorporate a database of 20 examples and an example selector to choose two examples that most closely resemble the given query. The process followed for few-shot learning is given in \cref{fig:fs-model}. Our custom example selector is based on cosine similarity to find the examples that most resemble the input. 

As for \textbf{Chain-of-Thought (CoT)} (\cref{fig:cot-model}), we guide the LLM in two steps: 

\textbf{Step 1:} Generation of colors and textures. We utilize few-shot learning and a template to incorporate the examples in the placeholders.

\textbf{Step 2:} Generation of outfit description. We exploit the colors and the textures generated to fill in the template presented in \cref{tab:templates} for CoT.

We perform \textbf{RAG} to inject knowledge to our model, utilizing two different data sources (see \cref{tab:sources}): PDF files of fashion articles and online pages of fashion blogs. The proposed template has two placeholders as shown in \cref{tab:templates}, one for the context and one for the question. In other words, we introduce an information retrieval component that harnesses user input to pull information from a new data source. In order to retrieve the most relevant information to the query, we use Langchain's vectorDB \cite{vectorDB} and then create embeddings. The user query and the relevant information are both given to the LLM.

\begin{table}[tb]
  \caption{Sources used for RAG}
  \label{tab:sources}
  \centering\small
  \begin{tabular}{@{}l@{}}
    \toprule
   Sources\\
    \midrule
    Encyclopedia of Clothing and Fashion \cite{Encyclopedia}\\
    Different Clothes for Different Occasions \cite{ClothesOccasions}\\
    A Study on Category of Female Body Shapes and their Clothing \cite{article}\\
    Fashion and Style Reference Guide \cite{FashionReferenceGuide}\\
  \bottomrule
  \end{tabular}
\end{table}

\begin{table}[]
  \caption{Prompt templates used for different prompting techniques}
  \label{tab:templates}
  \centering\small
  \begin{tabular}{@{}p{0.8cm} p{11cm}@{}}
    \toprule
     & Prompt template\\
    \midrule
    \textbf{ZS} & Imagine you are an expert in fashion design. Write a description for a fashion outfit in $[style]$ style appropriate for a $[type]$ at a $[occasion]$. Be sure to address the colors and the textures.\\ \midrule
    \textbf{FS} &  $[Examples]$ Question: Imagine you are an expert in fashion design. Write a description for a fashion outfit in $[style]$ style appropriate for a $[type]$ at a $[occasion]$. Be sure to address the colors and the textures. Answer: \\ \midrule
    \textbf{CoT} & Imagine you are an expert in fashion design. Write a description for a fashion outfit in $[style]$ style appropriate for a $[type]$ at a $[occasion]$. Be sure to use these colors: $[colors]$ and these textures: $[textures]$. \\ \midrule
    \textbf{RAG} & $[INST]<>$ Imagine you are a fashion expert. Always be creative and innovative. If the answer is not present in the context, make up one by yourself $<> CONTEXT: $ $[context]$$<> REQUEST: $ $[question][/INST]$\\ 
    \bottomrule
  \end{tabular}
\end{table}

\section{Experiments}

To evaluate our proposed method and compare various prompting/RAG techniques we carry out various experiments. These experiments are executed on T4 GPU, ensuring efficient processing without being computationally expensive.

\subsection{Dataset}
The dataset "fashion-style-instruct" \cite{Dataset} includes style recommendations based on input triplets of body type, personal clothing style, and event context, with GPT-3.5 \cite{openai2024chatgpt} generating outfit suggestions. We process the dataset to focus on \textit{body types} and \textit{occasions}, creating final triplets for two \textit{simple} types (differentiated by gender) and two \textit{complex} types (incorporating gender and body type). We consider 10 occasions ("a music festival", "a wedding", "a bachelor party", "a play/concert", "a job interview", "a business meeting", "a work/office event", "a tropical vacation", "a cruise") and 5 styles (gothic, classic, casual, bohemian, sporty). Our experiments combine these elements, producing 100 triplets for simple types and 100 for complex types.

\subsection{Evaluation}

\subsubsection*{CLIPscore} \cite{hessel2022clipscorereferencefreeevaluationmetric} evaluates text-image alignment by encoding text descriptions and images into fixed-length vectors and then computing the cosine similarity between these vectors. Using contrastive learning, CLIP \cite{DBLP:journals/corr/abs-2103-00020} is trained to bring similar text-image pairs closer and push dissimilar pairs apart in the embedding space. The resulting cosine similarity score, ranging from -1 (completely dissimilar) to 1 (completely similar), indicates the alignment strength between the text and image, with higher values signifying stronger alignment.

\subsubsection*{Human Evaluation} is an essential component of our method, since humans are the final receivers of the proposed outfits. In our conducted survey, participants are initially asked to provide demographic information to elucidate their background. Subsequently, they engage in three separate experiments. The survey was anonymous and voluntary, ensuring participant privacy and unbiased feedback. 

In the first experiment, a fashion outfit image is provided, and participants are asked to evaluate its visual appeal, relevance, creativity and overall impression. It includes the following questions:
\vspace{0.7em} 
\hrule
\begin{enumerate} \small
    \item \textit{On a scale of 1 to 5, how well does the outfit align with the style?}
    \item \textit{On a scale of 1 to 5, how suitable is the outfit for the occasion?}
    \item \textit{On a scale of 1 to 5, how fitting is the outfit for the type?}
    \item \textit{On a scale of 1 to 5, how creative is the outfit?}
    \item \textit{On a scale of 1 to 5, rate the aesthetic appeal of the outfit.}
    \item \textit{On a scale of 1 to 5, how well do the clothes and accessories match in the outfit?}
    \item \textit{Are there any abnormalities or inconsistencies in the image?}
    \item \textit{If the answer in the previous question was yes: Despite any abnormalities or inconsistencies, do you believe the image could serve as inspiration for a fashion designer?}
\end{enumerate}
\hrule
\vspace{0.7em} 
In the second experiment, participants are presented with a fashion outfit description and asked to assess its clarity, coherence, and relevance. It includes the following questions:
\vspace{0.7em} 
\hrule
\begin{enumerate}\small
    \item \textit{On a scale of 1 to 5, how comprehensible is the description?}
    \item \textit{On a scale of 1 to 5, how coherent is the description?}
    \item \textit{On a scale of 1 to 5, how suitable is the outfit described for the occasion?}
    \item \textit{On a scale of 1 to 5, how suitable is the outfit described for a type?}
    \item \textit{On a scale of 1 to 5, how well does the outfit described align with the style?}
    \item \textit{On a scale of 1 to 5, how suitable are the colors used for the occasion?}
    \item \textit{On a scale of 1 to 5, how suitable are the colors used for a type?}
    \item \textit{On a scale of 1 to 5, how suitable are the colors used for a style?}
    \item \textit{On a scale of 1 to 5, how suitable are the textures used for the occasion?}
    \item \textit{On a scale of 1 to 5, how suitable are the textures used for a type?}
    \item \textit{On a scale of 1 to 5, how suitable are the textures used for a style?}
\end{enumerate}
\hrule
\vspace{0.7em} 

In the third experiment, participants compare fashion outfits produced by the generative model with descriptions generated from 5 different techniques. 

\subsubsection*{Chi-square statistic} and \textbf{p-values} were used to assess statistical differences between the performance metrics of various methods. The chi-square test measures the discrepancy between observed and expected frequencies, indicating whether the observed performance differences are significant. A p-value of 0.05 was used as the significance threshold; p-values less than or equal to 0.05 suggest that the differences are unlikely due to chance, providing strong evidence against the null hypothesis and indicating genuine variations in method performance.

\section{Results}

\subsection{Overall quantitative performance}

To assess the performance of Stable Diffusion in conjunction to input prompts, we calculate CLIPscore to evaluate the generated image based on each prompting/RAG choice. The results, which demonstrate each method's effectiveness, range from 0.29-0.31, indicating medium text-image alignment. Apparently, all reported techniques are comparable for both simple and complex types. Thus, it is essential to analyze the human survey results to draw definitive conclusions.

\subsection{Human evaluation results}

\subsubsection*{Demographic analysis}

Participants in the study, totaling 79, are predominantly young adults with nearly equal representation of males and females. Their occupations vary, with a majority being students or full-time employed, but few are primarily involved in art-related fields, limiting expertise in fashion. Despite above-average interest in fashion, engagement in related activities (shopping, watching fashion shows etc) is infrequent. English proficiency among participants is generally high, with most being proficient and fewer at an intermediate level. In terms of survey experience, a significant portion have not participated in AI surveys previously, and even fewer have engaged in fashion-related surveys. 

\begin{figure}[h!]
\centering
\begin{subfigure}{.5\textwidth}
  \centering
   \includegraphics[width=.4\linewidth]{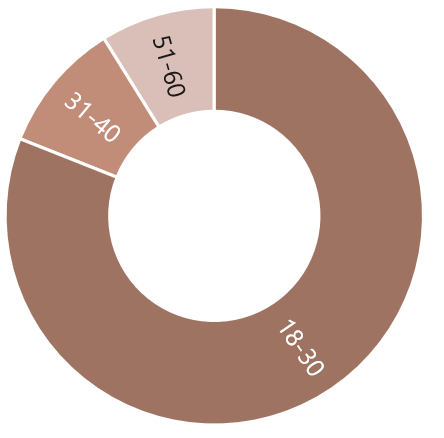}
   \caption{Age Range of participants}
\end{subfigure}%
\begin{subfigure}{.5\textwidth}
 \centering
 \includegraphics[width=.45\linewidth]{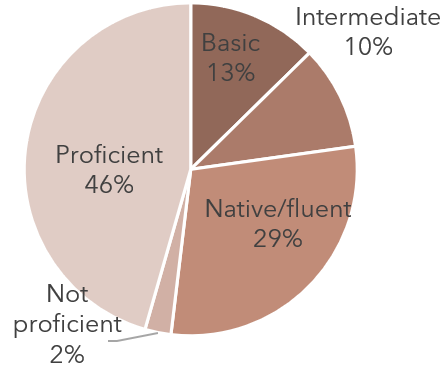}
 \caption{English level of participants}
\end{subfigure}
\begin{subfigure}{.5\textwidth}
  \centering
   \includegraphics[width=.65\linewidth]{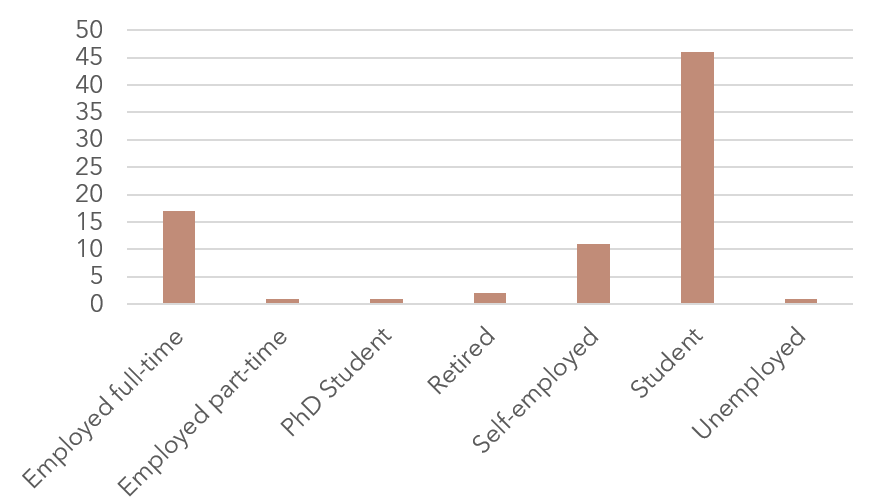}
   \caption{Primary occupation of participants}
\end{subfigure}%
\begin{subfigure}{.5\textwidth}
 \centering
 \includegraphics[width=.4\linewidth]{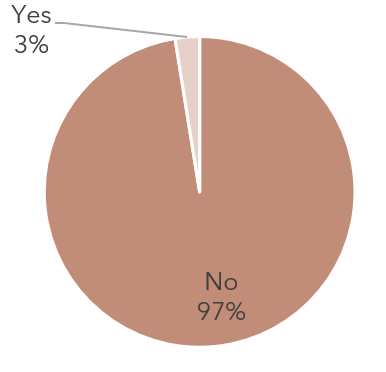}
 \caption{Relation of primary occupation with art}
\end{subfigure}
\begin{subfigure}{.5\textwidth}
  \centering
   \includegraphics[width=.4\linewidth]{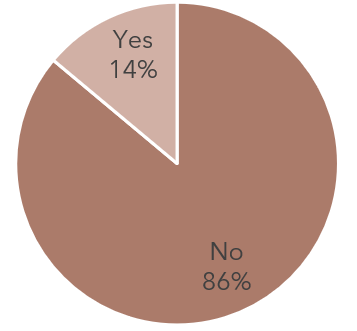}
   \caption{Participation in other fashion surveys}
\end{subfigure}%
\begin{subfigure}{.5\textwidth}
 \centering
 \includegraphics[width=.4\linewidth]{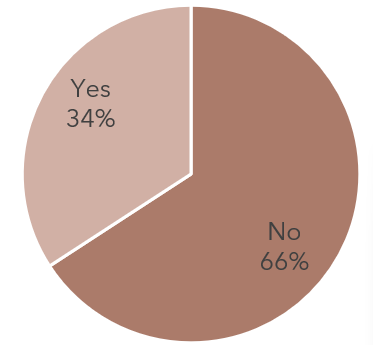}
 \caption{Participation in other fashion surveys}
\end{subfigure}
\label{fig:parti1}
\caption{Demographic profiles of survey participants.}
\end{figure}

\subsubsection{First Experiment}

\begin{figure}[h!]
\centering
\includegraphics[width=0.7\linewidth]{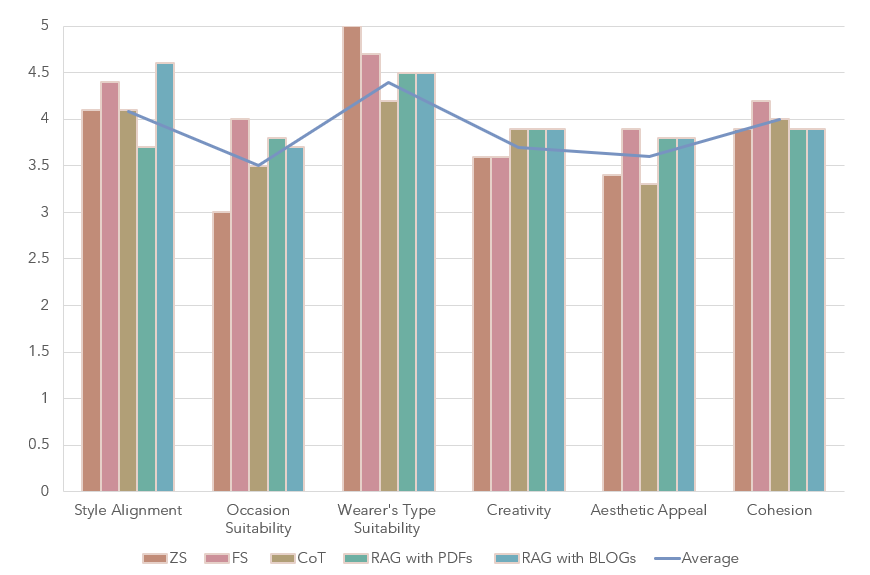}
\caption{Evaluation comparison of fashion images generated by descriptions of different techniques including ZS, FS, CoT, RAG with PDFs, and RAG with BLOGs across various criteria. Criteria considered are Style Alignment, Occasion Suitability, Wearer's Type Suitability, Creativity, Aesthetic Appeal, and Cohesion, with average performance indicated by the blue line.}
\label{fig:imageval}
\end{figure}

In the first experiment, participants are asked questions to gain a perspective about the generated images. As observed in Fig. \ref{fig:imageval}, participants rated the style of the outfits highly, with a mean rating of 4.1 out of 5, indicating strong alignment with the intended design. The suitability of the outfits for various occasions received a moderate rating of 3.5, suggesting some variability in participant opinions on this aspect.

However, there was a strong consensus (mean rating of 4.4) on how well the outfits matched the wearer's type, indicating a clear understanding and agreement among participants. In terms of creativity, aesthetic appeal, and coherence, the majority of participants rated the outfits as moderately to very creative. Additionally, participants generally agreed that the garments and accessories complemented each other well, contributing to a cohesive overall look.

\begin{figure}
\centering
\begin{subfigure}{.48\textwidth}
  \centering
   \includegraphics[width=.41\linewidth]{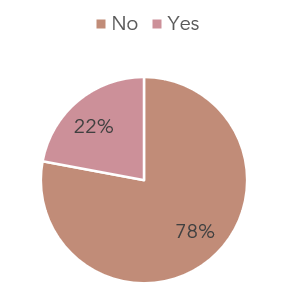}
   \caption{Percentage of images with and without abnormalities. \vspace{\baselineskip}}
   \label{fig:abn-gr}
\end{subfigure}%
\hfill
\begin{subfigure}{.48\textwidth}
 \centering
 \includegraphics[width=.45\linewidth]{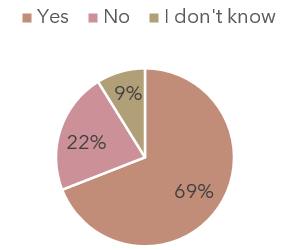}
 \caption{Distribution of responses regarding whether fashion designers find inspiration despite abnormalities.}
 \label{fig:abn2-gr}
\end{subfigure}
\label{fig:abnor}
\caption{Analysis of abnormalities in images and their influence on fashion designers' inspiration.}
\end{figure}

Regarding the quality of the generated images, 78\% of respondents indicated that they did not observe any abnormalities or inconsistencies, tied with Stable Diffusion's generation process, as shown in Fig. \ref{fig:abn-gr}. However, it is important to highlight that among those who noted inconsistencies, the overwhelming majority considered them to be not significant enough to detract from the outfit's potential as a source of inspiration for fashion designers, as shown in Fig. \ref{fig:abn2-gr}.

Having gained a comprehensive understanding of the generated results, it is beneficial to delve into the different responses from participants to images generated using descriptions produced by various methods. This examination can provide deeper insights into how different descriptive approaches influence participant perception, identifying related patterns and preferences.

As shown in Fig. \ref{fig:imageval}, descriptions generated through FS learning performed better than ZS in most cases. Regarding outfit - style alignment, FS had a statistical difference of  $\chi^2$ = 27.58, $p$ = \(3.5 \times 10^-3\) with ZS  and it gained first place among the other methods regarding outfit-occasion alignment. 
When measuring the creativity of the outfit, participants favored methods including RAG and CoT. Especially CoT outperformed ZS with $\chi^2$ = 25.87, $p$ = \(56 \times 10^-3\).
FS and RAG with PDFs and BLOGs were also voted highly for producing descriptions resulting in aesthetically pleasing images. More specifically, FS compared to ZS had $\chi^2$ = 27.99, $p$ = \(32 \times 10^-3\).
Regarding the coherence of the outfit (how much the clothes and the accessories match), FS was voted better than the other methods surpassing the second in order CoT and presenting a statistical difference of $\chi^2$ = 35.48, $p$ = \(34 \times 10^-4\) with each other.

\subsubsection{Second Experiment}

While evaluating the quality of generated images through participant feedback offers valuable insights, a more insightful approach involves analyzing their responses to the generated descriptions directly. This method allows for a thorough assessment of how effectively the language model conveys fashion concepts and characteristics through textual descriptions.

\begin{figure}[h!]
\centering
\includegraphics[width=0.9\linewidth]{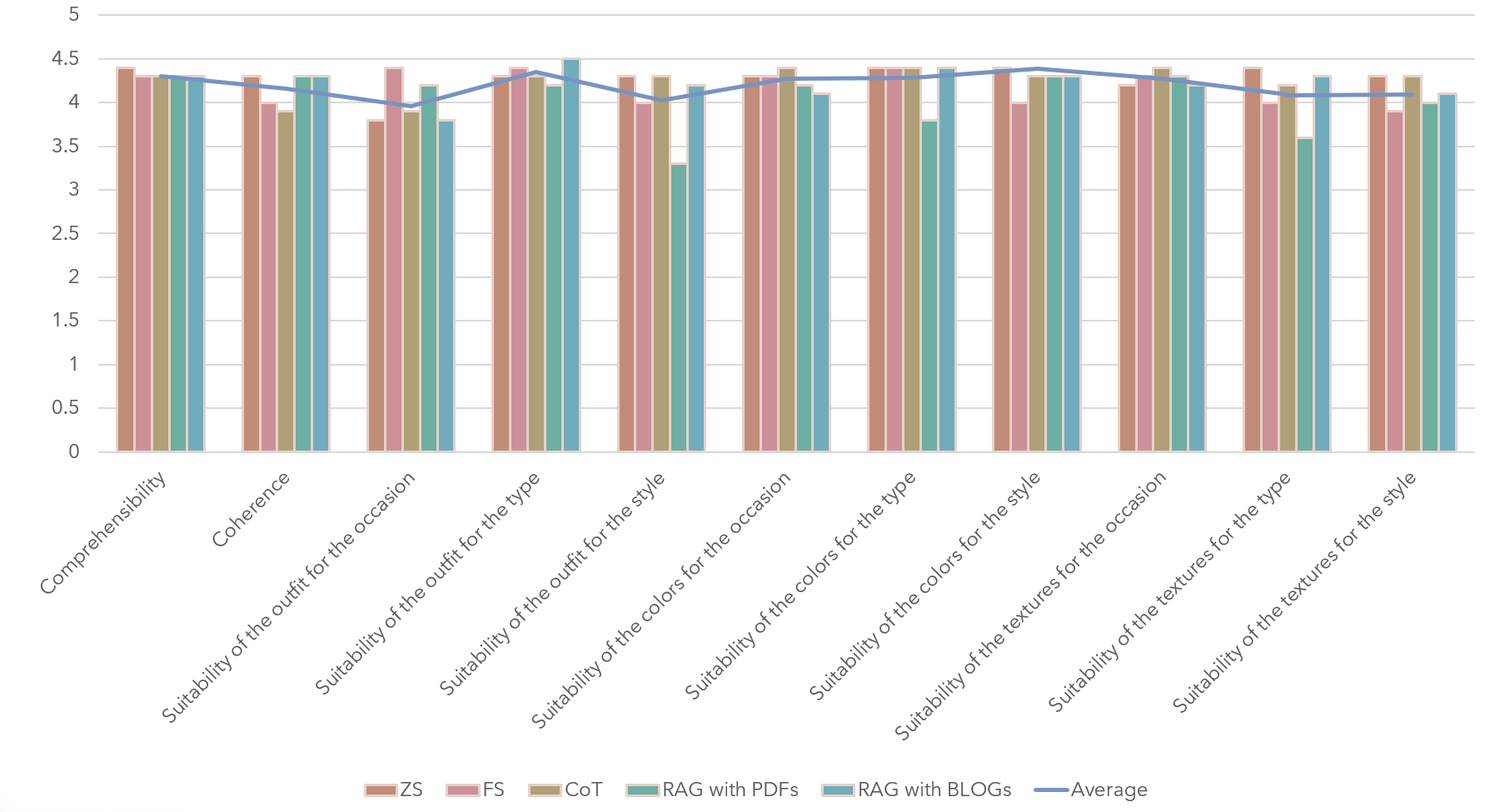}
\caption{Evaluation comparison of descriptions generated by different techniques  including ZS, FS, CoT, RAG with PDFs, and RAG with BLOGs across various criteria.}
\label{fig:descrcomp}
\end{figure}

As shown in Fig. \ref{fig:descrcomp}, participants rated the comprehensibility and cohesion of the generated descriptions highly, with both aspects receiving an average score of $\sim$ 4.3. Notably, participants who identified as proficient or native/fluent in English provided overall higher scores compared to the rest. This distinction underscores the importance of language proficiency in the evaluation process and suggests that the descriptions were generally well-crafted.

In addition to comprehensibility and cohesion, participants also rated the outfit descriptions based on the suitability of the outfit for the type, the style, and the occasion. The generated outfits performed well across these variables. Notably, the suitability for the occasion received a particularly high score of almost 4.5. This score is significantly higher than the ratings given for the alignment of the outfits presented in the images with the occasion. This comparison highlights the effectiveness of the descriptions in conveying appropriate outfits for various occasions, suggesting that the textual descriptions were more successful in meeting participants' expectations than the visual representations. Participants found the descriptions highly suitable for the type, indicating a strong match with the described body types. Similarly, the suitability for the style received strong ratings, showing alignment with the intended fashion styles. The colors described were also well-received, with participants considering them appropriate for both the occasion and the type, reflecting thoughtful and appropriate color selection. Moreover, the suitability of colors for the style was rated positively, demonstrating harmony between the color choices and the fashion styles described.
When it comes to textures, the ratings were equally high. Participants felt that the textures described were suitable for the occasion. The textures were also considered well-matched with the type and the style and type of the wearer.

Among the top performers, RAG with BLOGs and PDFs consistently achieved high scores across most criteria. Similarly, CoT and FS  performed admirably; CoT particularly excelled in coherence, suggesting that step-by-step reasoning can enhance the logical flow of the descriptions. FS showed strength in several areas, making it a robust option for generating high-quality descriptions.
However, there are areas for improvement, particularly for ZS, which generally lagged behind the others, especially in "Suitability of the colors for the occasion" and "Suitability of the outfit for the type." The lower scores suggest that ZS might struggle with specific contextual nuances, indicating a need for further refinement to improve its performance in these areas.

All methods performed relatively well in terms of coherence and comprehensibility, though variations were observed. ZS and RAGs were noted for their high rating in coherence with great statistical differences.

Each method offers specific insights. More specifically, when it comes to outfit-occasion alignment FS was rated higher with its p-value and chi-square statistics in relevance to the second best RAG with PDFs ($\chi^2$ = 35.42, $p$ = \(4 \times 10^-3\)). The latter showed great statistical difference with its superior methods in the field of outfit-occasion alignment, having $\chi^2$ = 46.5 , $p$ = \(10^-3\) with ZS, and $\chi^2$ = 41.71, $p$ = \(4.2 \times 10^-3\) with RAG with BLOGS.
In relevance to outfit-wearer's type alignment, participants gave the best average score to RAG with BLOGS followed by FS with narrow statistical difference of $\chi^2$ = 9.6, $p$ = 0.38 between them. RAG with BLOGs however excelled when compared with ZS ($\chi^2$ = 57.41, $p$ = \(10^-8\)) and CoT ($\chi^2$ = 36.48, $p$ = \(3 \times 10^-5\)).
Concerning the suitability of the outfit for the given style, RAG with BLOGs received a higher rating and presented a significant statistical difference with RAG with PDFs ($\chi^2$ = 36.44, $p$ = \(3 \times 10^-3\)). It is also interesting to note that ZS outperformed FS with $\chi^2$ = 59.17, $p$ = \(10^-7\).
Regarding the question relevant to the alignment of colors with the occasion, CoT was rated as  superior having $\chi^2$ = 24.85, $p$ = \(15.6 \times 10^-3\) with RAG with BLOGs. Promising results were given by the rating of ZS and FS which outperformed RAG with BLOGs with $\chi^2$ = 57.56, $p$ = \(10^-6\) and $\chi^2$ = 37.53, $p$ = \(10^-7\) accordingly.  ZS presents a higher rating as the suitability of the colors for a given style, as demonstrated by the metrics $\chi^2$ = 32.67 $p$ = \(1.1 \times 10^-3\) with RAG with BLOGs and $\chi^2$ = 36.6, $p$ = \(6 \times 10^-4\) with FS.
Investigating the correlation between texture and occasion, CoT is rated higher than the others, showing $\chi^2$ = 23.92 $p$ = 0.02 with FS which was second in ranking, having same average with RAG with BLOGs. FS performed better in this aspect compared to ZS according to the participants with a statistical difference of $\chi^2$ = 34.52 $p$ = 0.0006. Moreover, in this case RAG with BLOGs results' achieved greater scores than RAG with PDFs with $\chi^2$ = 41.58 $p$ = \(4.6 \times 10^-4\).
As for the questions about which textures are best for different wearers' types ZS achieved impressive ratings by passing RAG with BLOGs with  $\chi^2$ = 38.8 $p$ =\(10^-6\). RAG with BLOGs achieve the second best average score surpassing CoT with $\chi^2$ = 45.09 $p$ = \(1.3 \times 10^-4\) and FS with $\chi^2$ = 95.32 $p$ = \(10^-13\).
Lastly, for the texture-style alignment ZS and CoT gained the first place among the methods with  $\chi^2$ and $p$ values indicating great statistical difference in the distributions between ZS and FS, ZS and RAG (either with PDFs or with BLOGs) and between CoT and RAGs.

Overall, the evaluation highlights the strengths and areas for improvement of different methods used in generating fashion outfit descriptions. RAG with PDFs and FS emerge as the most effective approaches, providing comprehensive and contextually relevant descriptions. RAG with BLOGs and CoT also show strong potential. On the other hand, ZS, while promising, needs further refinement to improve its contextual understanding.

\subsubsection*{Comparison of LLMs results}
We evaluate and compare the outputs of the Falcon and Mistral models based on participant feedback from the second experiment. The study analyzes user preferences regarding the descriptions generated by these models to determine which model produces outputs perceived as more accurate, fluent, and contextually appropriate by human evaluators.

\begin{figure}[h!]
    \centering \small
    \begin{subfigure}{0.3\textwidth}
        \centering
        \includegraphics[width=\linewidth]{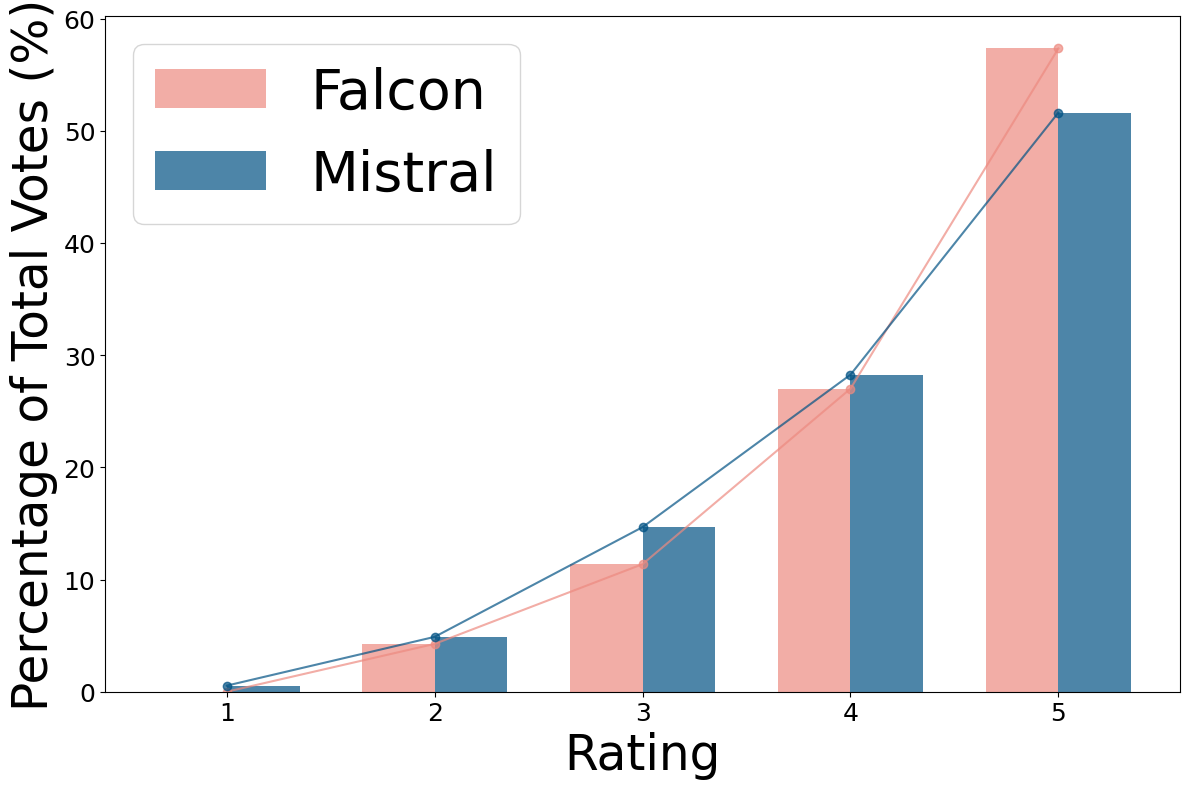}
        \caption{Description comprehensibility}
        \label{fig:sub1}
    \end{subfigure}
    \hfill
    \begin{subfigure}{0.3\textwidth}
        \centering
        \includegraphics[width=\linewidth]{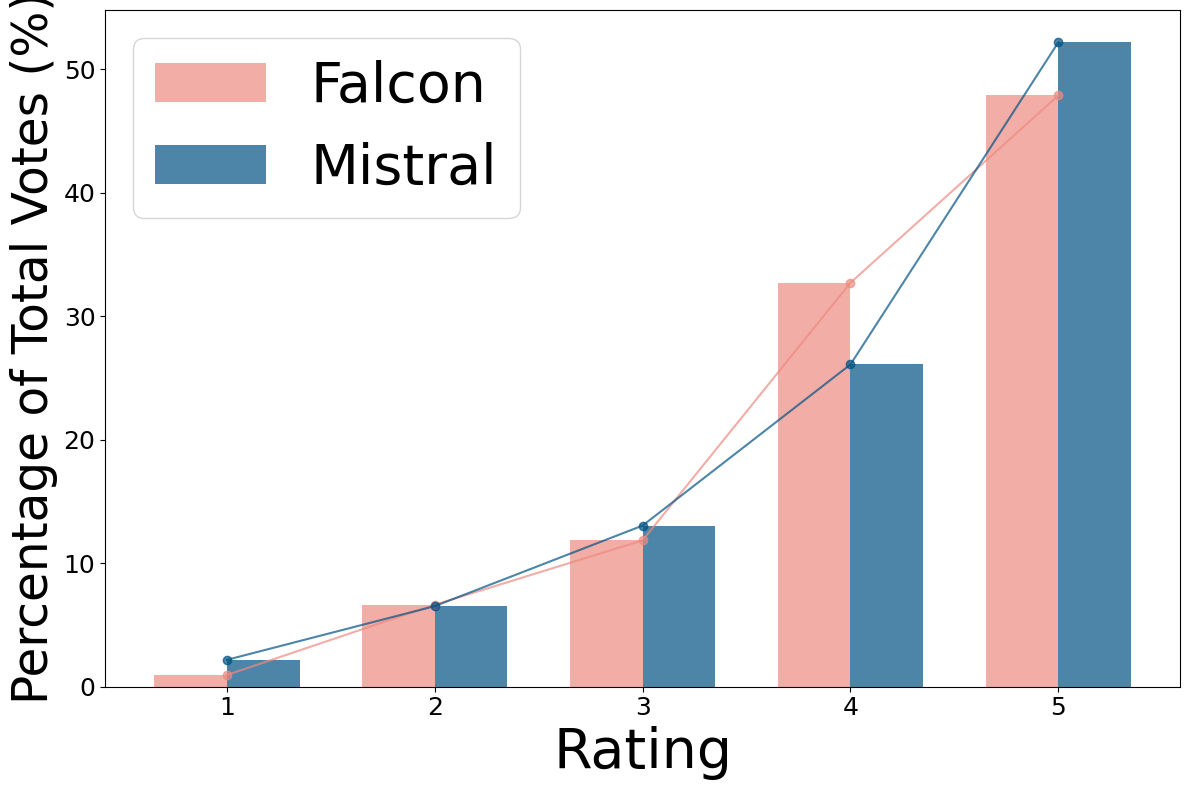}
        \caption{Description coherence  \vspace{\baselineskip}}
        \label{fig:sub2}
    \end{subfigure}
    \hfill
    \begin{subfigure}{0.3\textwidth}
        \centering
        \includegraphics[width=\linewidth]{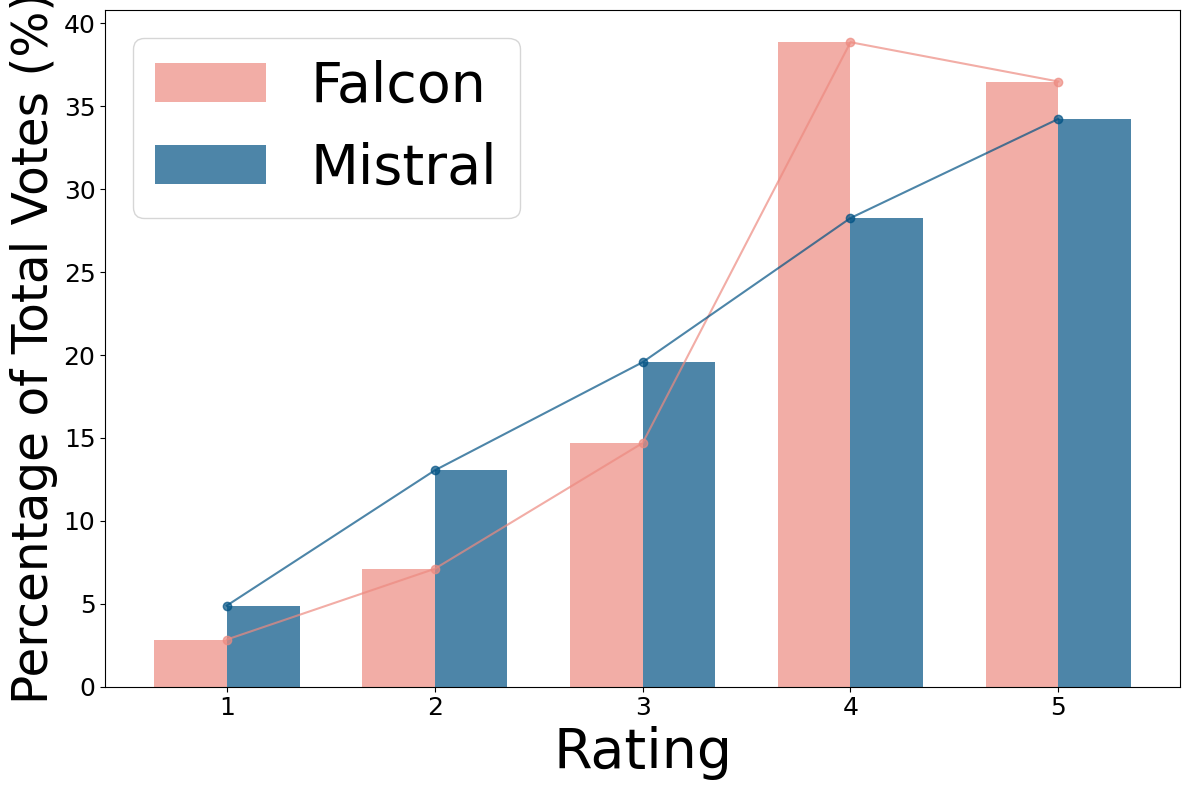}
        \caption{Outfit-occasion alignment \smallskip}
        \label{fig:sub3}
    \end{subfigure} 
    \hfill
     \begin{subfigure}{0.3\textwidth}
        \centering
        \includegraphics[width=\linewidth]{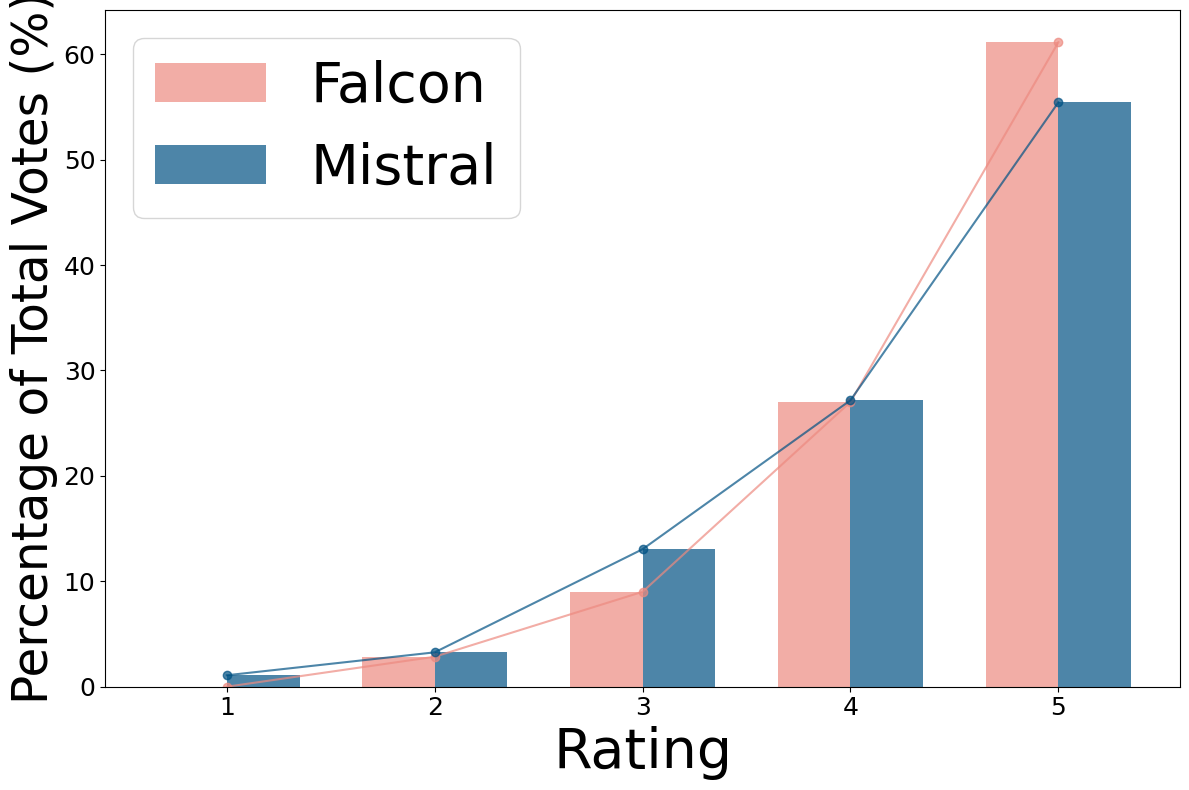}
        \caption{Outfit-wearer's type alignment}
        \label{fig:sub4}
    \end{subfigure}  
    \hfill
    \begin{subfigure}{0.3\textwidth}
        \centering
        \includegraphics[width=\linewidth]{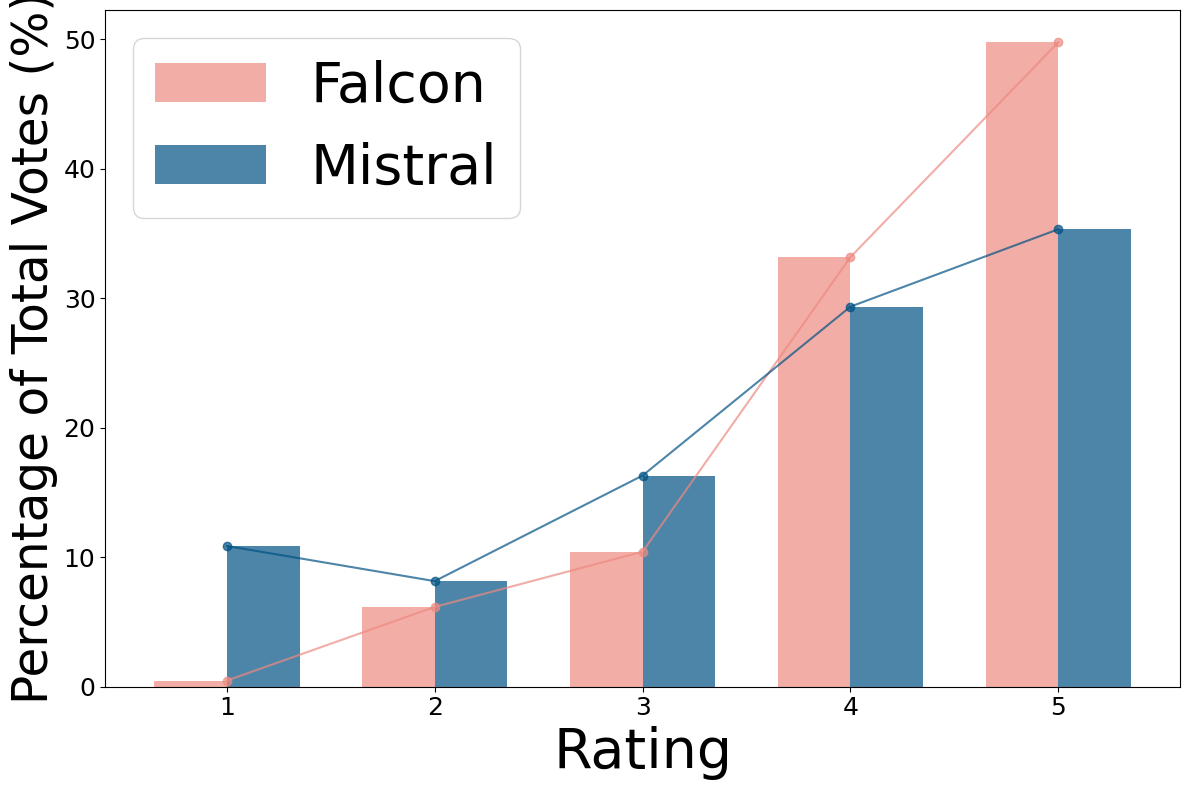}
        \caption{Outfit-style alignment\vspace{\baselineskip}}
        \label{fig:sub5}
    \end{subfigure} 
        \hfill
       \begin{subfigure}{0.3\textwidth}
        \centering
        \includegraphics[width=\linewidth]{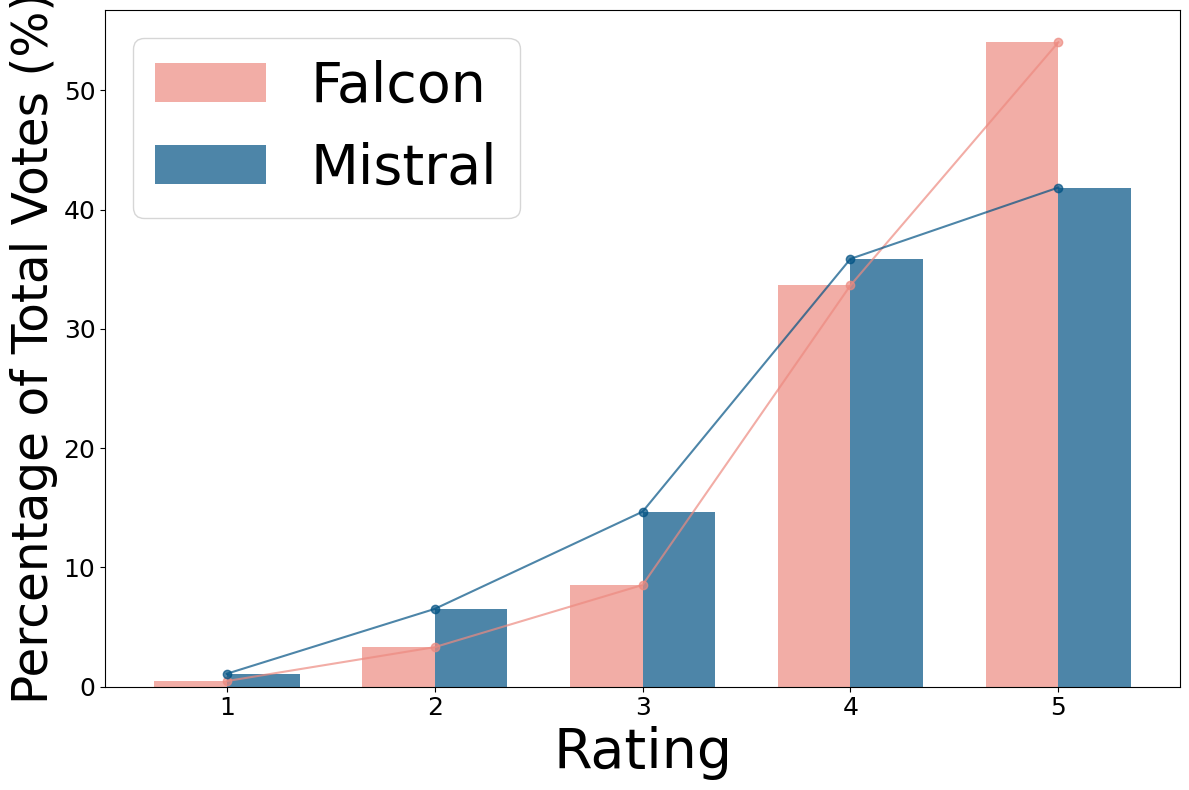}
        \caption{Color-occassion alignment\vspace{\baselineskip}}
        \label{fig:sub6}
    \end{subfigure} 
     \hfill
     \begin{subfigure}{0.3\textwidth}
        \centering
        \includegraphics[width=\linewidth]{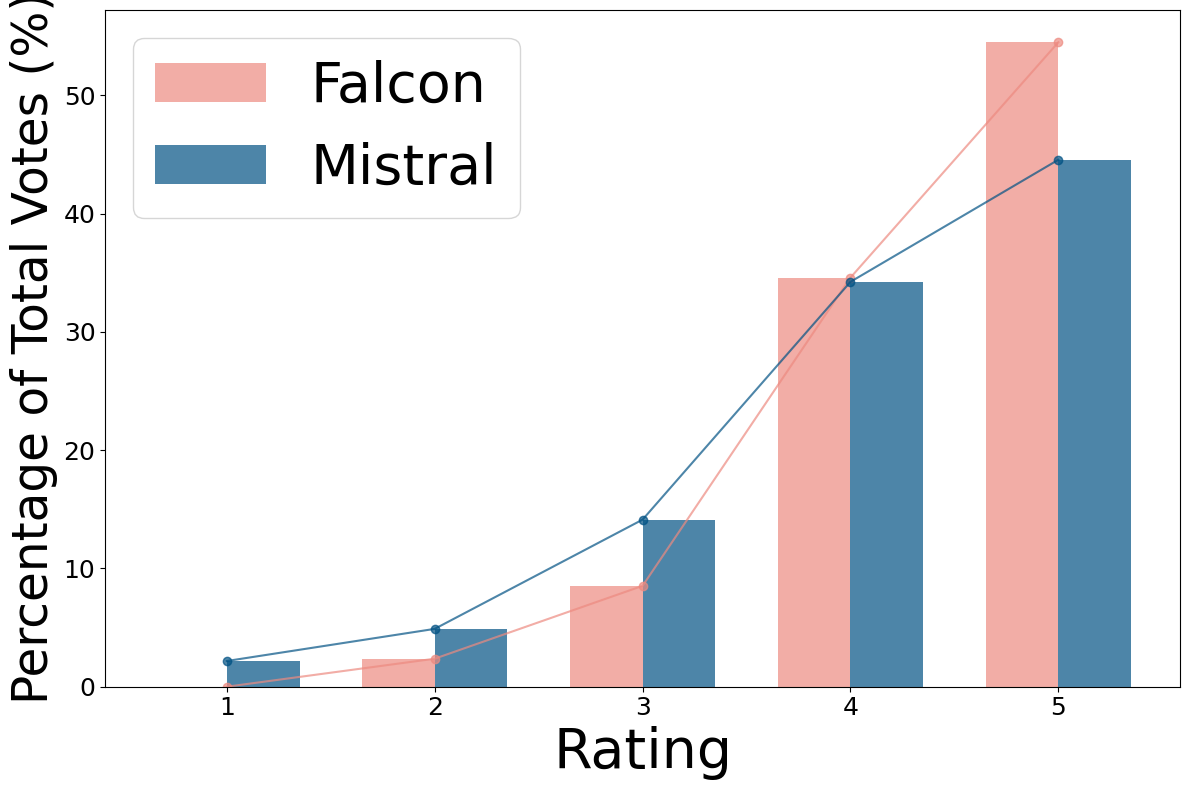}
        \caption{Color-wearer's type alignment}
        \label{fig:sub7}
    \end{subfigure} 
    \hfill
    \begin{subfigure}{0.3\textwidth}
        \centering
        \includegraphics[width=\linewidth]{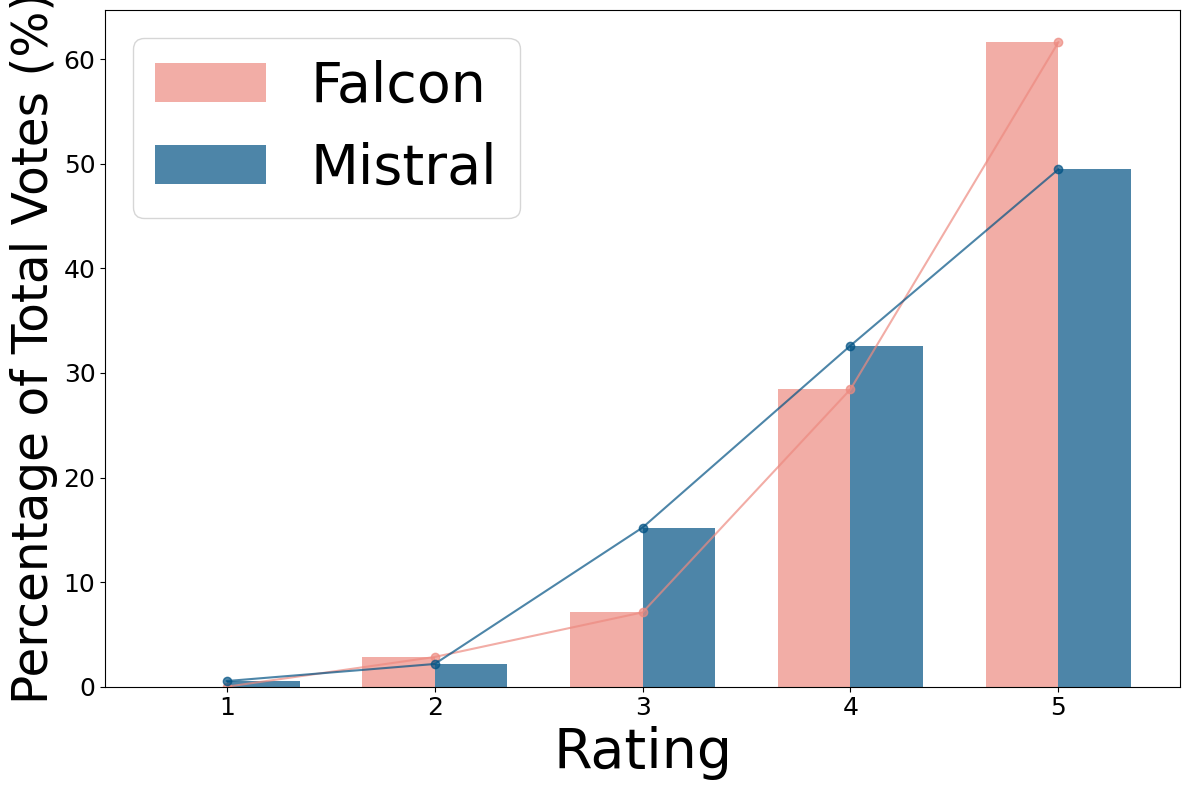}
        \caption{Color-style alignment \vspace{\baselineskip}}
        \label{fig:sub8}
    \end{subfigure} 
    \hfill
       \begin{subfigure}{0.3\textwidth}
        \centering
        \includegraphics[width=\linewidth]{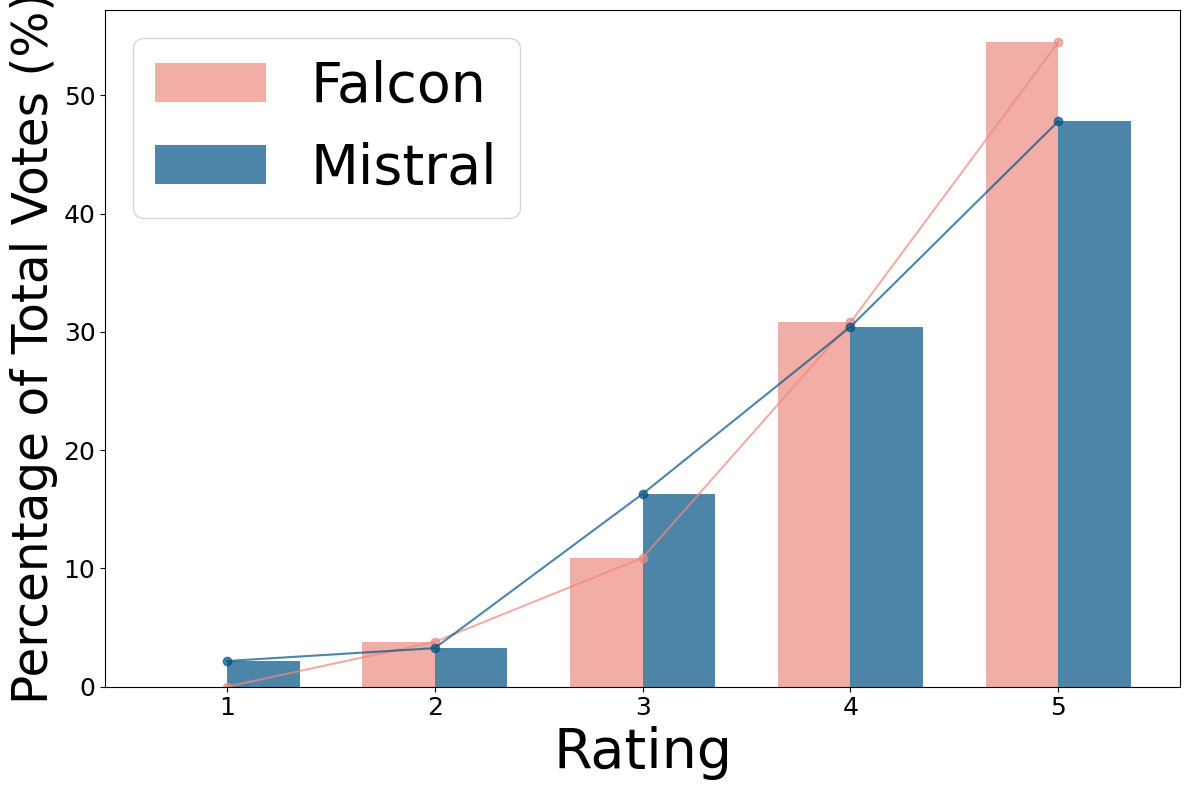}
        \caption{Texture-occasion alignment}
        \label{fig:sub9}
    \end{subfigure} 
    \hfill
     \begin{subfigure}{0.3\textwidth}
        \centering
        \includegraphics[width=\linewidth]{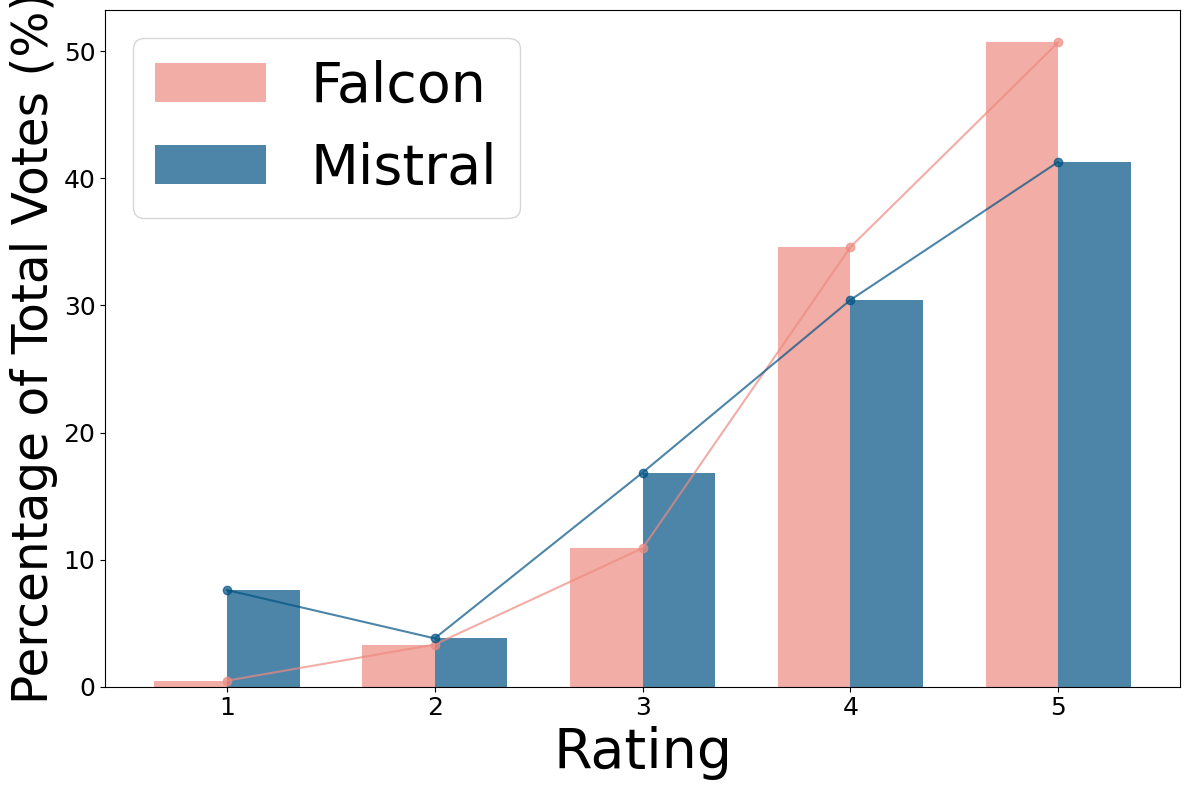}
        \caption{Texture-wearer's type alignment}
        \label{fig:sub10}
    \end{subfigure}
    \hspace{0.6cm}
    \begin{subfigure}{0.3\textwidth}
        \centering
        \includegraphics[width=\linewidth]{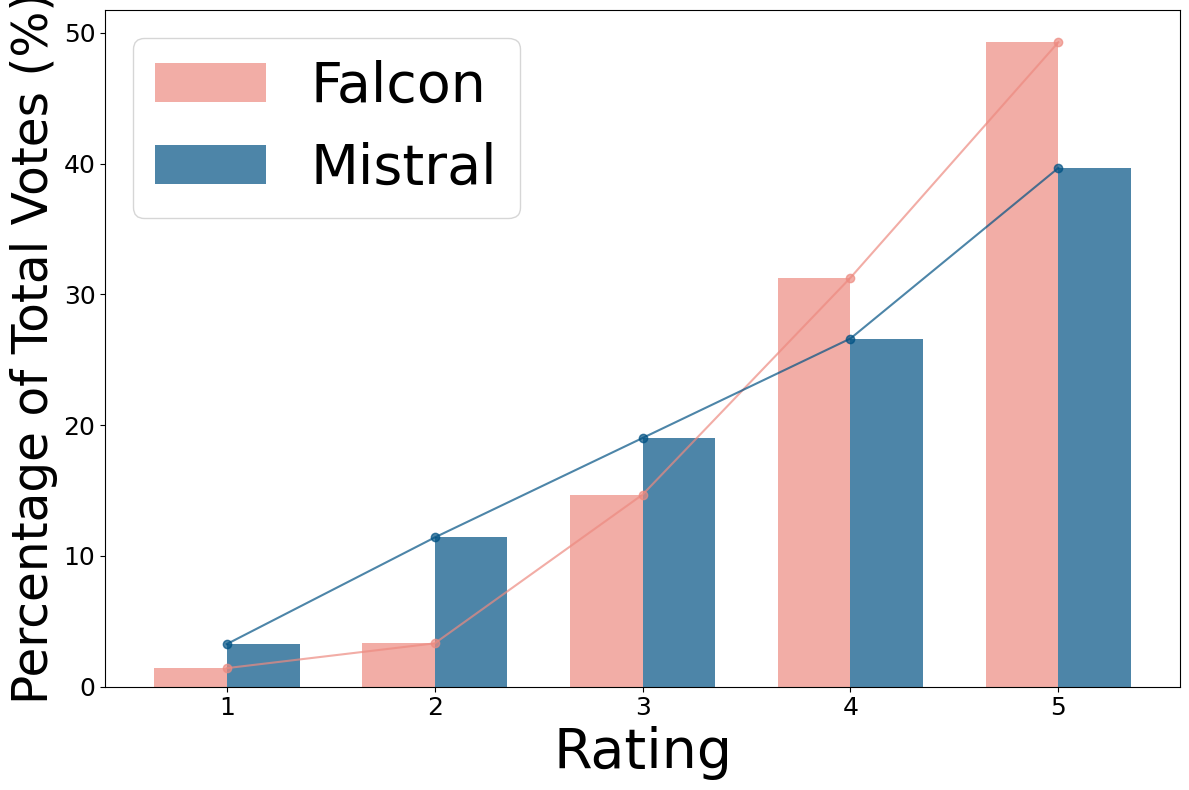}
        \caption{Texture-style alignment\vspace{\baselineskip}}
        \label{fig:sub11}
    \end{subfigure} 
    \caption{Comparison of human evaluation results for the two LLMs.}
    \label{fig:main}
\end{figure}

In each subfigure of \cref{fig:main}, the horizontal axis represents the grades given by participants to the outfit descriptions, ranging from 1 (lowest) to 5 (highest). The vertical axis shows the percentage of participants assigning each grade. The criterion used by participants to assess the descriptions is specified in the caption of each figure. In the comparative analysis, Falcon consistently outperformed Mistral in several key criteria. Falcon received higher scores for comprehensibility, indicating clearer descriptions. Mistral, however, showed superiority in coherence, with better logical flow and consistency. Falcon significantly excelled in aligning outfits with the occasion and the wearer's type, demonstrating better contextual understanding and personalization capabilities. Falcon also led in aligning outfits with specified styles, colors, and textures, consistently achieving higher scores across these metrics. Overall, while Mistral was strong in clarity and coherence, Falcon stood out for its contextual relevance and personalized fashion recommendations.

\subsubsection*{Third Experiment}
\begin{figure}[tb]
\centering
\includegraphics[width=0.6\linewidth]{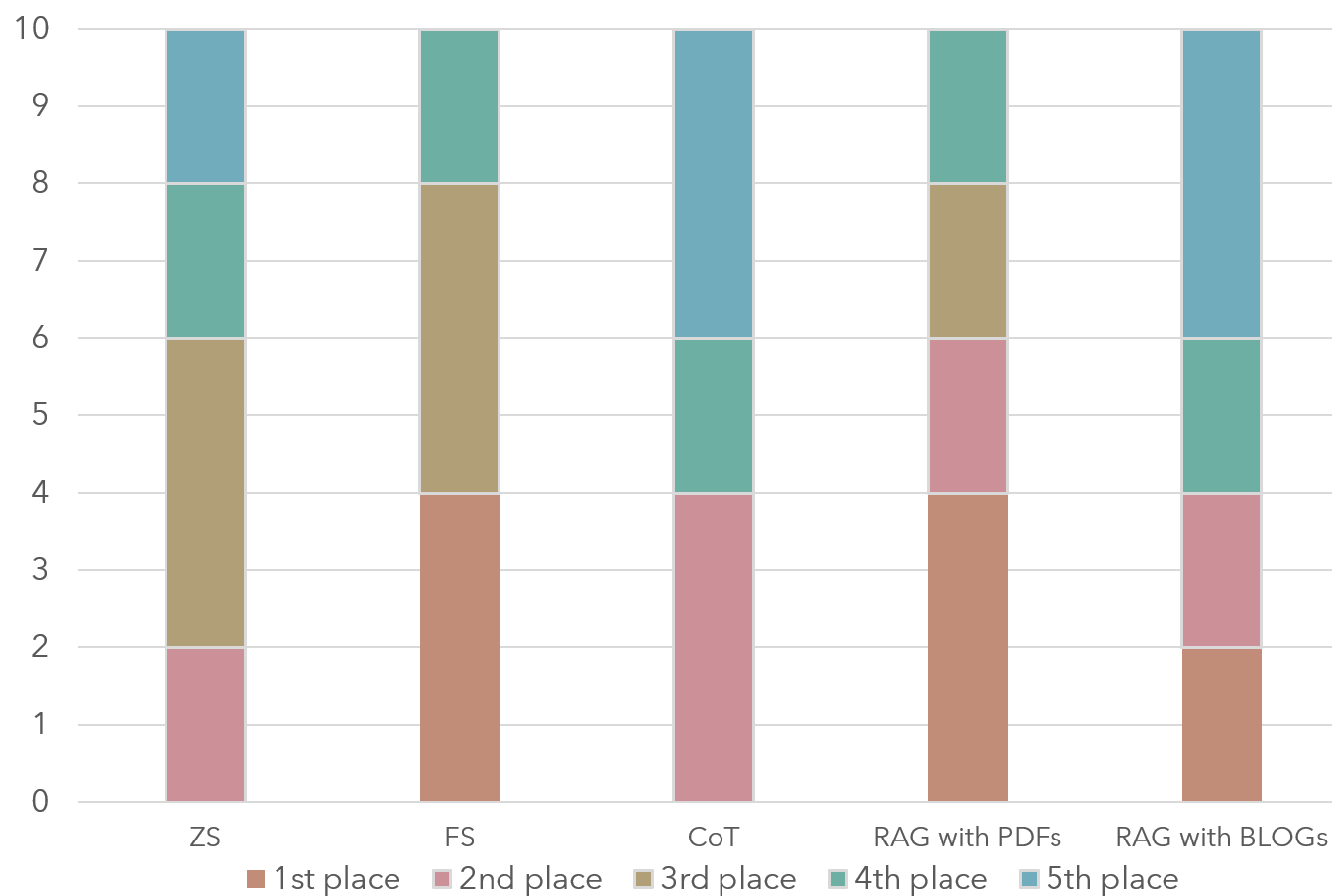}
\caption{\begin{minipage}{0.9\linewidth}
\centering Distribution of comparative ranks for prompting/RAG according to humans.
\end{minipage}}
\label{fig:winners}
\end{figure}

In this experiment, participants are asked to choose their preferred outfit from images generated using descriptions from the five different prompting/RAG methods, comparing the aesthetic appeal of images in relation to the occasion, style, and wearer's type, focusing on the LLM input used.
\cref{fig:winners} displays the distribution of rankings (1st-5th place) across the five methods. FS appears to be the most consistently high-performing method, while the others exhibit more variability across the different rankings. ZS and RAG  show balanced performance but are less likely to achieve the top rank compared to FS. 

\subsection{Qualitative Results}

\begin{figure}
\centering
\includegraphics[width=1\linewidth]{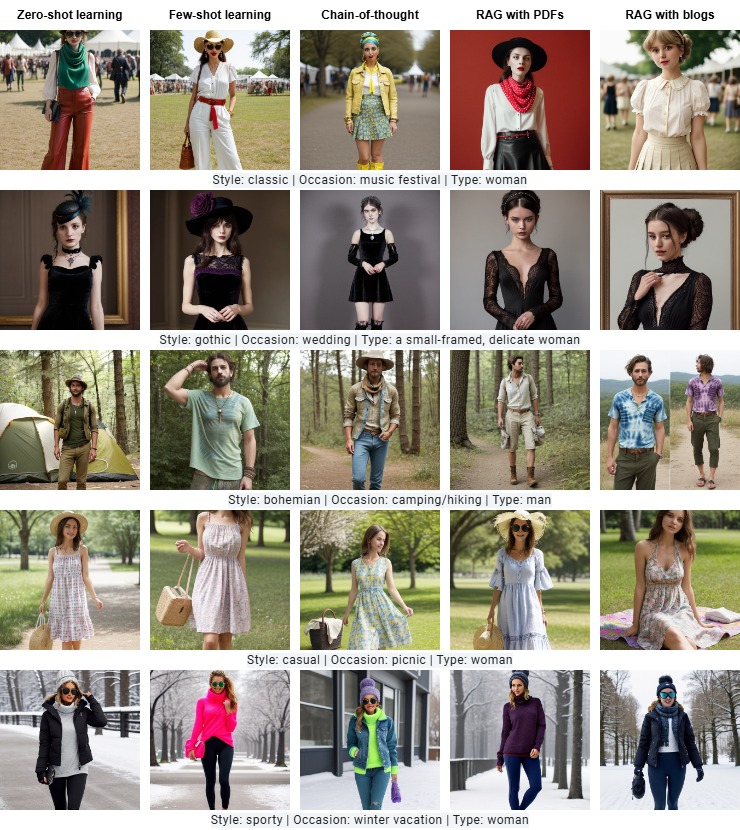}
\caption{Images generated with Stable Diffusion when given descriptions of 5 different methods}
\label{fig:myres1}
\end{figure}

The visual results are illustrated in \cref{fig:myres1}. Backgrounds, such as parks, snowy landscapes, and crowds, provide contextual clues to suggest occasions like picnics, winter vacations, and festivals, enhancing the realism and richness of the visuals. Some images offer multiple garment, color, or texture options for the same variable set, demonstrating the model's flexibility but also introducing potential ambiguity that may need refinement for more clear and consistent outputs.

\section{Conclusion}

This study investigates generating fashion descriptions using two LLMs and a Stable Diffusion model for creating images, focusing on a prompting approach. The RAG method, which integrates insights from sources like fashion magazines and blogs, enhances the models' trend awareness. Evaluations using quantitative metrics and human judgment revealed that the generated descriptions were highly aligned, creative, coherent, and aesthetically appealing. Future research should extend RAG to include social media, fashion archives, and user-generated content to boost trend diversity and adaptability. Additionally, incorporating varied prompts and fashion variables could yield more personalized fashion outputs, while using ontologies or scene graphs could enhance the semantic understanding and coherence of the results.

%
%
\bibliographystyle{splncs04}
\bibliography{main}
\end{document}